\documentclass[letterpaper, 10 pt, conference]{ieeeconf}  
\usepackage{booktabs}
\usepackage{nccmath, mathtools}
\usepackage{graphicx}   
\usepackage{xcolor}
\usepackage{subcaption}
\usepackage{wrapfig,lipsum}
\usepackage{amsmath,amssymb,amsfonts}
\usepackage{multirow}
\usepackage{cite}
\usepackage[font=footnotesize,labelfont=bf]{caption}
\usepackage{nomencl}

\IEEEoverridecommandlockouts                              

\title{\LARGE \bf
Bigraph Matching Weighted with Learnt Incentive Function for Multi-Robot Task Allocation
}

\author{Steve Paul$^{1}$, Nathan Maurer$^{2}$ and Souma Chowdhury$^{3,\dagger}$
\thanks{$^\dagger$ Corresponding Author, soumacho@buffalo.edu}
\thanks{Authors $^{1}, ^{2}, ^{3}$ are with the Department of Mechanical and Aerospace Engineering, University at Buffalo, Buffalo, NY, USA {\tt\small \{stevepau, namaurer, soumacho\}@buffalo.edu}}
\thanks{This work was supported by the Office of Naval Research (ONR) award N00014-21-1-2530 and the National Science Foundation (NSF) award CMMI 2048020. Any opinions, findings, conclusions, or recommendations expressed in this paper are those of the authors and do not necessarily reflect the views of the ONR or the NSF.}
}

\begin{document}

\maketitle
\thispagestyle{empty}
\pagestyle{empty}

\begin{abstract}
Most real-world Multi-Robot Task Allocation (MRTA) problems require fast and efficient decision-making, which is often achieved using heuristics-aided methods such as genetic algorithms, auction-based methods, and bipartite graph matching methods. These methods often assume a form that lends better explainability compared to an end-to-end (learnt) neural network based policy for MRTA. However, deriving suitable heuristics can be tedious, risky and in some cases impractical if problems are too complex. This raises the question: can these heuristics be learned? To this end, this paper particularly develops a Graph Reinforcement Learning (GRL) framework to learn the heuristics or incentives for a bipartite graph matching approach to MRTA. Specifically a Capsule Attention policy model is used to learn how to weight task/robot pairings (edges) in the bipartite graph that connects the set of tasks to the set of robots. The original capsule attention network architecture is fundamentally modified by adding encoding of robots' state graph, and two Multihead Attention based decoders whose output are used to construct a LogNormal distribution matrix from which positive bigraph weights can be drawn. The performance of this new bigraph matching approach augmented with a GRL-derived incentive is found to be at par with the original bigraph matching approach that used expert-specified heuristics, with the former offering notable robustness benefits. During training, the learned incentive policy is found to get initially closer to the expert-specified incentive and then slightly deviate from its trend.
\end{abstract}

\section{Introduction}
\vspace{-.2cm}
Multi-Robot Task Allocation (MRTA) is the problem of allocating multiple robots to complete multiple tasks with the goal of maximizing/minimizing an objective or cost.  In this paper, we are interested in comparing learning methods that generalize well with more traditional optimization methods. Some of the real-world applications include construction \cite{meng2008distributed}, disaster response \cite{ghassemi2021multi}, manufacturing \cite{8785546}, and warehouse logistics \cite{xue2019task}. Even though expensive solvers such as Mixed-Integer Non-Linear Programming (MINLP) provide near-optimal solutions, these methods cannot be deployed for scenarios with 100s or 1000s of tasks and robots.

There exists a notable body of work on heuristic-based approaches to solving MRTA problems more efficiently, e.g., using genetic algorithm \cite{su13020902}, graph-based methods \cite{farinelli2008decentralised,Ghassemi-bigmrta-mrs2019,ghassemi2021multi}, and auction-based methods \cite{dias2006market}). Often the heuristic choices and setting in these approaches are driven by expert experience and intuition. Hence, although they provide some degree of explainability, they leave significant scope for improvement in performance. Moreover, heuristic-based methods are often poor at adapting to complex problem characteristics, or generalizing across different problem scenarios, without tedious hand-crafting of underlying heuristics. As a result, an emerging notion in (fast, near-optimal) decision-making is ``can more effective heuristics be automatically (machine) learned from prior operational data or experience gathered by the agent/robot e.g., in simulation" \cite{mazyavkina2021reinforcement}? This raises the fundamental question of whether learned heuristics can match and potentially surpass the performance of human-prescribed heuristics when generalizing across a wide range of problem scenario of similar or varying complexity. We explore this research question in the context of multi-robot task allocation problems of the following type Multi Robot Tasks-Single Task Robots (MR-ST) \cite{paul2023efficient}. In doing so, this paper also provides initial evidence of the potential for exploiting the best of both worlds: explainable structure of graph matching techniques and automated generation of the necessary heuristics through reinforcement learning (RL).





\vspace{-.1cm}
\subsection{Related Works}\label{subsec:Related_works}
\vspace{-.2cm}
Some of the most common online solution approaches for MRTA problems include heuristic-based methods Integer linear programming (ILP) based methods, bipartite graph (bigraph) matching, meta-heuristic methods such as genetic algorithm and Ant Colony Optimization, and also Reinforcement Learning (RL) based methods.
ILP-based mTSP formulations and solutions have also been extended to MRTA problems~\cite{jose2016task}. Although the ILP-based approaches can in theory provide optimal solutions, they are characterized by exploding computational effort as the number of robots and tasks increases~\cite{toth2014vehicle,cattaruzza2016vehicle}. 

Most online MRTA methods, e.g., auction-based methods~\cite{dias2006market,schneider2015auction,otte2020auctions}, metaheuristic methods \cite{Mitiche2019, VANSTEENWEGEN20093281}, and bi-graph matching methods~\cite{ismail2017decentralized,ghassemi2021multi,farinelli2008decentralised}, Genetic Algorithms \cite{su13020902}, use some sort of heuristics, and often report the optimality gap at least for smaller test cases compared to the exact ILP solutions. \cite{farinelli2008decentralised} introduces the use of bigraphs and the Max-Sum algorithm for decentralized task allocation in multi-agent systems. specifically, maximum weighted bipartite matching \cite{ghassemi2021multi,farinelli2008decentralised} with manually tuned incentive functions (aka expert heuristics) has been shown to provide scalable and effective solutions to various MRTA problems. The incentive function typically represents the affinity of any given robot to choose a given task based on the task's properties and the state of that robot. However every time the features of the robot, robot team, or task space changes, the incentive functions must be re-designed or re-calibrated by an expert to preserve performance. This technical challenge motivates learning of the incentive function from experience.

Over the past few years, Graph Reinforcement Learning (GRL) methods have emerged as a promising approach to solving combinatorial optimization problems encountered in single and multi-agent planning applications \cite{Kool2019, Kaempfer2018LearningTM, khalil2017learning, paul2022scalable, Tolstaya2020MultiRobotCA, Paul_ICRA, 9750805, paulaviation2022, paul2023efficient, krisshnakumar2023fast, kumar2023graph}. While these methods have shown to generate solutions that can generalize across scenarios drawn from the same distribution and can be executed near instantaneously, they are considered to be black-box methods (thus lacking explainability) and usually do not provide any sort of guarantees. 

\textbf{Key Contributions:} The overall objective of this paper is to identify an approach to learning the incentive function that can be used by maximum weighted bigraph matching to perform multi-robot task allocation, with performance that is comparable to or better than reported i) bigraph matching techniques that use expert heuristics, and ii) purely RL based solutions. Thus the main contributions of this paper include: \textbf{1)} Identify the inputs, outputs and structure of the graph neural network (GNN) model that will serve as the incentive function; \textbf{2)} Integrate the GNN-based incentive with the bigraph matching process in a way that the GNN can be trained by policy gradient approaches over simulated MRTA experiences; \textbf{3)} Analyze the (dis)similarity of the learned incentives to that of the expert-derived incentives.

\textbf{Paper Outline:} The next section summarizes the MRTA problem, its MDP formulation, and the bigraph representation of the MRTA process. Section \ref{sec:Learning_Framework} describes our proposed GNN architecture for incentive learning and computing the final action (robot/task allocations). Section \ref{sec:Experimental_Evaluations} discusses numerical experiments on MRTA problems of different sizes, comparing learning and non-learning methods and analyzing computing time. Section \ref{sec:Conclusion} concludes the paper.
\vspace{-.2cm}
\section{MRTA - Collective Transport (MRTA-CT)}\vspace{-.2cm}
\label{sec:Multi_Robot_collective_Transport}


Here we consider the MRTA Collective Transport (MRTA-CT) problem defined in \cite{paul2023efficient}. Given a homogeneous set of $N^{R}$ robots denoted as $R$ (${r_{1},r_{2},\ldots,r_{N^{R}}}$) and a set of $N^{T}$ tasks represented by $V^{T}$, the objective is to optimize task allocation to maximize the number of completed tasks. There is a central depot that serves as both the starting and ending points for each robot. Each task, denoted as $i \in V^{T}$, possesses unique location coordinates, $(x_{i}, y_{i})$. Additionally, each task has a workload or demand, $w_{i}$, which may vary over time, and a time deadline, $\tau_i$, by which the task must be completed (i.e., demand satisfied) to be considered as ``done'' ($\psi=1$). Each robot has a maximum range, $\Delta_{\text{max}}$ that limits its total travel distance including return to depot. Robots also have a predefined maximum payload carrying capacity, $C_{\text{max}}$. A robot starts its journey from the depot with a full battery and payload, proceeds to task locations to partly or completely fulfill its demands; it returns to the depot once it's either completely unloaded, running low on battery, or there are no remaining tasks in the environment, whichever condition is met first. The recharging process is assumed to be instantaneous, such as through a battery swap mechanism.

\vspace{-.1cm}
\subsection{MRTA-CT as Optimization Problem}\label{sec:mrta_optimization}
\vspace{-.1cm}

The exact solution to the MRTA-CT problem is obtained by formulating it as an INLP problem, as concisely expressed below (for brevity); details can be found in \cite{paul2023efficient}.
\begin{align}\small
\vspace{-.5cm}
\label{eq:objectiveFunction}
 \min~ f_\text{cost} &= (N^{T} - N_\text{success})/{N^{T}} \\ 
   N_{\text{success}} &= \sum_{i \in V^{T}} \psi_{i} \nonumber
    \begin{cases}
     \psi_{i} = 1, \ if \ \tau^{f}_{i} \leq \tau_{i} \\
     \psi_{i} = 0, \ if \ \tau^{f}_{i} > \tau_{i} 
    \end{cases}
\end{align}\vspace{-.6cm}
\begin{align}
\label{eq:constraint}
& 0 \leq \Delta_{r}^{t} \leq \Delta_{\text{max}},  r \in R \\
\label{eq:constrain2}
& 0 \leq c_{r}^{t} \leq C_{\text{max}} ,  r \in R
\vspace{-.3cm}
\end{align}
Here $\tau_{i}^{f}$ is the time at which task $i$ is completed, $\Delta_{r}^{t}$ is the available range for robot $r$ at a time instant $t$, 
$c_{r}^{t}$ is the capacity of robot $r$ at time $t$, $N_\text{success}$ is the number of successfully completed tasks during the operation. We craft the objective function (Eq.~\eqref{eq:objectiveFunction}) such that it emphasizes minimizing the rate of incomplete tasks, i.e., the number of incomplete tasks divided by the total number of tasks. Equations \ref{eq:constraint} and \ref{eq:constrain2} correspond to the remaining range and capacity respectively at time $t$.
We express the MRTA-CT as an MDP over a graph to learn policies that along with a bigraph matching approach yield solutions to this combinatorial optimization problem, as described next. 

\vspace{-.1cm}
\subsection{Bipartite Graph for MRTA}\label{subsec:BIGMRTA}
\vspace{-.1cm}

A bipartite graph (bigraph) is a graph with two distinct and non-overlapping vertex sets such that every edge in the graph connects a vertex from one set to a vertex in the other set. A weighted bigraph is a special type of bigraph in which each edge is assigned a numerical weight or cost. These weights represent some measure of significance, incentive, affinity, cost, or strength associated with the connections between the vertices in the graph.
\begin{wrapfigure}[14]{r}{0.2\textwidth}
\vspace{-0.2cm}
\scriptsize
    \centering
    \includegraphics[width=0.2\textwidth]{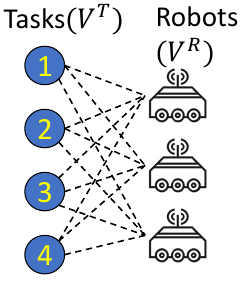}
    \vspace{-0.6cm}
    \caption{Bigraph showing robot-task connections. The Bigraph weights is written as a matrix.}
    \label{fig:bigraph}
    \vspace{-.4cm}
\end{wrapfigure}
Weighted bigraphs are commonly used to model and solve various real-world problems, where the weights on edges provide additional information about the relationships between entities across the two sets. In this work as shown in Fig. \ref{fig:bigraph}, we construct a bigraph $\mathcal{G}^{B} = (V^{B}, E^{B}, \Omega^{B})$, where the two types of vertices in $V^{B}$ are from the set of robots, $V^{R}$, and the set of tasks, $V^{T}$. Here, $E^{B}$ represent the edges that connect any vertex or node in $V^{R}$ with a vertex in $V^{T}$. Here, $\Omega^{B} (\in \mathbb{R}^{N^{R} \times N^{T}})$ is the weight matrix, where each weight, $\Omega^{B}_{r,i}, r \in V^{R}, i \in V^{T}$, is associated with the edge that connects nodes $r$ and $i$. In our MRTA representation, this weight $\Omega^{B}_{r,i}, r \in V^{R}, i \in V^{T}$ provides a measure of the affinity for robot $r$ to perform task $i$.

\textbf{\textit{Maximum Weight Matching}:} Weighted (bigraph) graph matching is a well-studied problem in graph theory and combinatorial optimization. The goal is to find a matching (a set of non-overlapping edges or one-to-one connections) in a bigraph such that the sum of the weights of the selected edges is maximized or minimized, depending on whether it's a maximum-weight or minimum-weight matching, respectively. Here we make use of maximum-weight matching to allocate tasks to the robots. Popular methods for weight matching for bigraphs include the Hungarian Algorithm \cite{kuhn1955hungarian} and Karp Algorithm \cite{hopcroft1973n}. Along with providing provably optimal matching, 
Maximum weight matching offers clarity and transparency on how robots are paired with tasks in the case of MRTA. Effectiveness of this approach however hinges on how well the weight matrix represent the \textit{relative} affinity or value of the robots to select tasks (connected by edges) given the current state of the environment. The formulation of the MDP to learn how to generate this weight matrix given the environment state is described next. 
\vspace{-.2cm}

\subsection{MDP over a Graph}\label{sec:ProblemFormulation}
\vspace{-.1cm}

The MDP is defined in an asynchronous decentralized manner for each individual robot, to capture its task selection process, and is expressed as a tuple, $<\mathcal{S}, \mathcal{A}, \mathcal{P}_a, \mathcal{R}>$. Here, we assume full observability, i.e., every robot communicates its chosen task with other robots.

\textbf{Graph formulation for Tasks:} The task space of an MRTA-CT problem is represented as a fully connected graph $\mathcal{G}^{T} = (V^{T}, E^{T}, \Omega^{T})$, where $V^{T}$ is a set of nodes representing the tasks, $E^{T}$ is the set of edges that connect the nodes to each other, and $\Omega^{T}$ is the weighted adjacency matrix that represents $E^{T}$. 
Node $i$ is assigned a 4-dimensional normalized feature vector denoting the task location coordinates, time deadline, and the remaining workload/demand i.e., $\delta^{T}_i$=$[x_i,y_i,\tau_i,w^t_i]$ where $i \in [1,N^{T}]$. Here, the edge weight between two nodes $\Omega^{T}_{i,j}$ ($\in \Omega^{T}$) is computed as $\Omega^{T}_{i,j} = 1 / (1+ |\delta^{T}_i - \delta^{T}_j|)$, where $i,j \in [1,N^{T}]$, and expresses how similar the two nodes are in the graph space. This is a common approach to compute the weighted adjacency matrix, despite the node properties representing different physical quantities. The degree matrix $D^{T}$ is a diagonal matrix with elements $D^{T}_{i,i} = \sum_{j \in V^{T}}\Omega^{T}_{i,j}, \forall i \in V^{T}$. 

\vspace{-.1cm}
\subsection{Task selection}
\vspace{-.1cm}

\textbf{\textit{Graph formulation for Robots}:} The state of the robots in MRTA-CT is represented as a fully connected graph $\mathcal{G}^{R} = (V^{R}, E^{R}, \Omega^{R})$, where $V^{R}$ is a set of nodes representing the robots, $E^{R}$ represent the set of edges, and $\Omega^{R}$ is the weighted adjacency matrix that represents $E^{R}$. The number of nodes and edges are $N^{R}$ and $N^{R}(N^{R}-1)$, respectively. Every robot node is defined as $\delta^{R}_{r} = [x^{t}_{r}, y^{t}_{r}, \Delta^{t}_{r},c^{t}_{r},t^{\text{next}}_{r}], \forall r \in R$, where for robot $r$, $x^{t}_{r}, y^{t}_{r}$ represents its current destination, $\Delta^{t}_{r}$ represents its remaining range (battery state), $c^{t}_{r}$ represents its remaining capacity of robot, and $t^{\text{next}}_{r}$ represents its next decision time. The weights of $\Omega^{R}$ are computed using $\Omega^{R}_{r,s}$ = $1 / (1+ |\delta^{R}_r - \delta^{R}_s|)$, where $r,s \in [1,N^{R}]$. The degree matrix $D^{R}$ is a diagonal matrix with elements $D^{R}_{r,r} = \sum_{s \in V^{R}}\Omega^{R}_{r,s}, \forall r \in V^{R}$. When a robot $r$ visits a task $i$, the demand fulfilled by the robot $r$ is ${\min}(w^t_i, c^{t}_{r})$.

The components of the MDP are defined as follows: \textbf{State Space ($\mathcal{S}$)}: a robot $r$ at its decision-making instance uses a state $s\in\mathcal{S}$, which contains the task graph $\mathcal{G}^{T}$ and the robots' state graph $\mathcal{G}^{R}$. 
\noindent\textbf{Action Space ($\mathcal{A}$):} each action $a$ is defined as the index of the selected task, $\{0,\ldots,N^{T}\}$ with the index of the depot as $0$. This action is selected as a result of the maximum weight matching of the bigraph. \noindent\textbf{State Transition, $\mathcal{P}_a(s'|s,a)$:} the transition is an event-based trigger. A robot taking action $a$ at state $s$ reaches the next state $s'$ in a deterministic manner. This definition assumes that the policy model encapsulates all processes, including but not limited to the learning-derived models, that together produces the action to be taken, given the state of the environment. \textbf{Reward ($\mathcal{R}$):} Here a reward of $1/N^{T}$ is given during each decision-making instance, if an active task (whose demand has not yet been fully met, and deadline has not passed) is chosen, while a reward of 0 is given if the depot is chosen. Hence, the maximum possible aggregate reward in any episode is 1.

\vspace{-.1cm}
\section{Incentive (Weight) learning framework}\label{sec:Learning_Framework}\vspace{-.2cm}

In our solution approach, we construct a policy network that serves as the incentive generator. Namely, it takes in the state information during a decision-making instance for a robot and outputs the bigraph weight matrix. This is used by maximum weight matching on the bigraph to yield the task selection for the deciding robot. This section discusses the policy network and the various steps involved in the sequential decision-making process.
\begin{figure}[!ht]
\vspace{-.35cm}
\scriptsize
    \centering
    \includegraphics[width=0.5\textwidth, height=0.18\textheight]{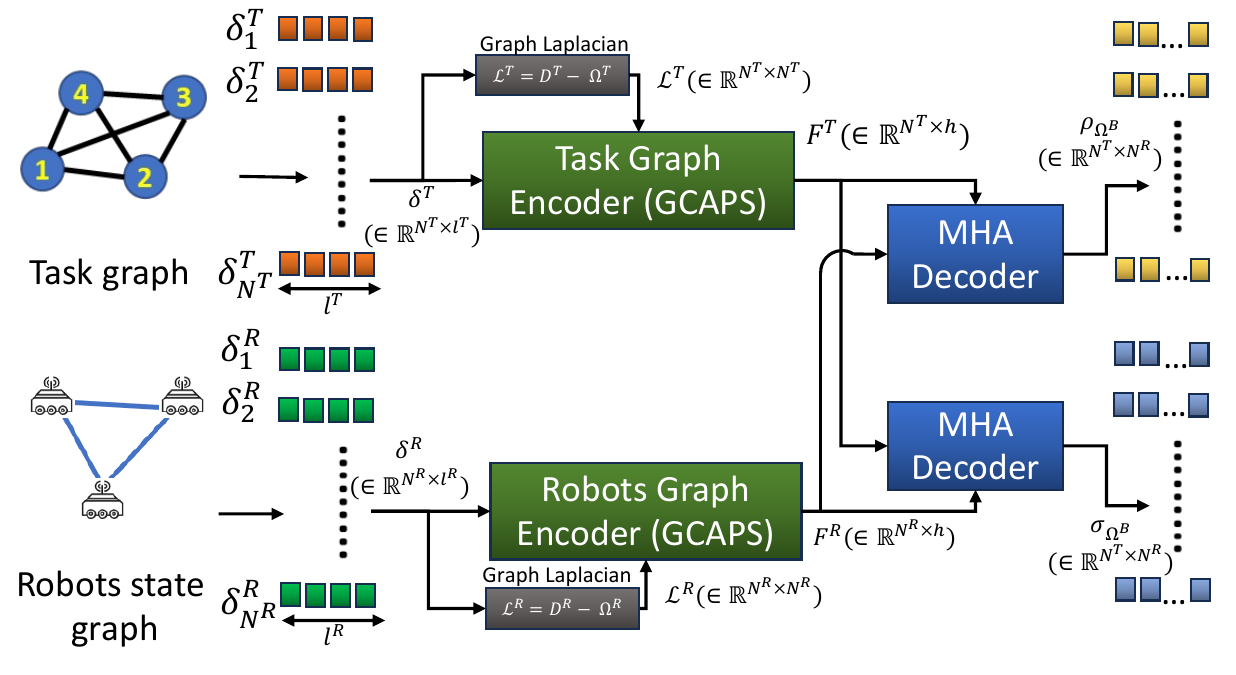}\vspace{-0.2cm}
    \caption{The overall structure of the BiG-CAM policy.}
    \label{fig:policy}
    \vspace{-0.3cm}
\end{figure}

Figure \ref{fig:policy} shows the inputs/outputs and structure of the policy model that computes the bigraph weights.
The elements of the weight matrix ($\Omega^{B}_{r,i}, r \in V^{R}, i \in V^{T}$), which represents an incentive score for that robot $r$ to pick task $i$ at that instance, should be computed as a function of the task and robot features. Here the policy model is designed to output parameters of (independent LogNormal) probability distributions from which the bigraph weights, $\Omega^{B}_{r,i}$, can be sampled. Since the state and features of the tasks and robots are both formulated as graphs ($\mathcal{G}^{T}$ and $\mathcal{G}^{R}$), we use GNNs to compute their node embeddings. Along with node features, the choice of the GNN type seeks to also embed the structure of the task and robot spaces to promote generalizability. These node embeddings are then used to compute the weights using Multi-head Attention-based decoders (MHA). We use two such decoders to produce a matrix of the mean values of the bigraph weights and a matrix of the corresponding standard deviation values. We call this incentive policy model \textbf{\textit{BiGraph-informing Capsule Attention Mechanism}} or \textbf{\textit{BiG-CAM}}, whose components are described next.

\begin{figure*}[!ht]
\begin{minipage}{0.49\linewidth}
\scriptsize
    \centering
    \includegraphics[width=1.0\textwidth, height=0.5\textwidth]{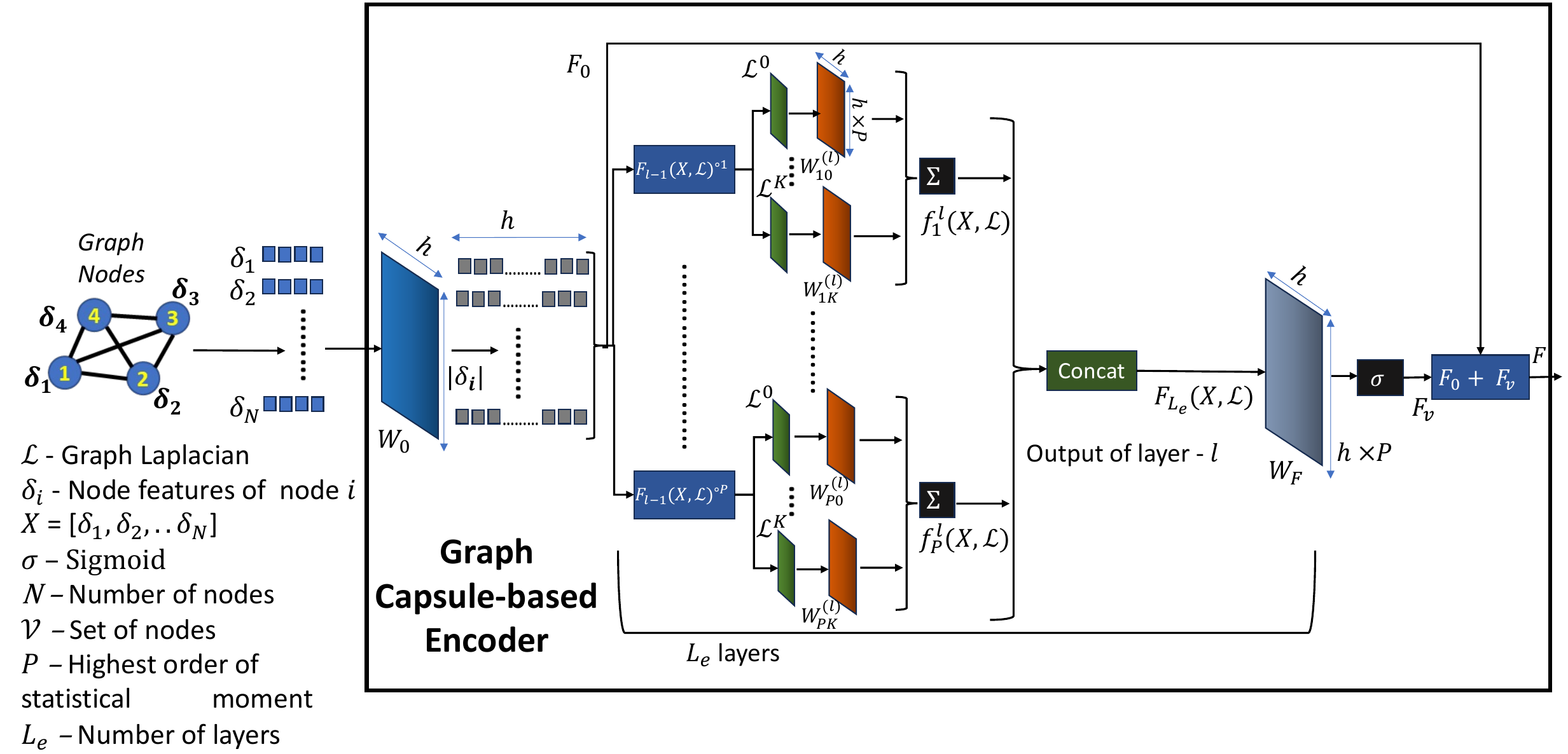}
    \caption{The overall structure of the GCAPS network. Here, $h$ is the embedding length, and bias terms are omitted for ease of representation.}
    \label{fig:gcaps}
\end{minipage}
\begin{minipage}{0.49\linewidth}
\scriptsize
    \centering
    \includegraphics[width=1.0\textwidth, , height=0.4\textwidth]{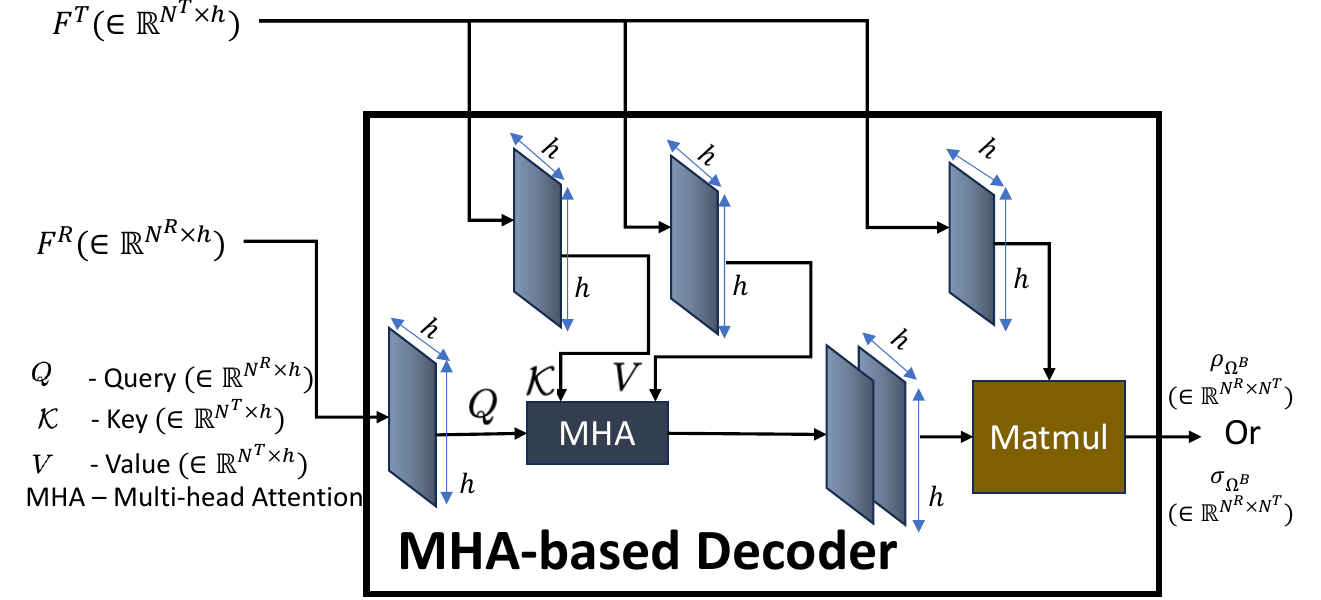}
    \caption{Structure of the MHA-based decoder.}
    \label{fig:mha_decoder}
\end{minipage}
\vspace{-.9cm}
\end{figure*}
\vspace{-.2cm}
\subsection{GNN-based feature encoder}\label{subsec:GNN-abstraction}

We use Graph Neural Networks (GNNs) for encoding the state of the tasks and robots. The GNNs take in a graph and compute node embeddings. We consider two separate encoders (Fig. \ref{fig:policy}) for the task and robot graphs, named as the Task Graph Encoder (TGE) and the Robots Graph Encoder (RGE). Both TGE and RGE are based on Graph Capsule Convolutional Neural Networks (GCAPS) \cite{Verma2018}, as shown in Fig. \ref{fig:gcaps}. In our prior works \cite{Paul_ICRA, paul2022scalable, paul2023efficient}, GCAPS has shown superior capability in capturing structural information compared to GNNs such as Graph Convolutional Networks (GCN) and Graph Attention Networks (GAT). Here, both the TGE and the RGE take in the corresponding graphs, $\mathcal{G}^{T}$ and $\mathcal{G}^{R}$, and compute the corresponding node embeddings, $F^{T}$ ($\mathbb{R}^{N^{T}\times h}$) and $F^{R}$ ($\mathbb{R}^{N^{R}\times h}$), respectively; here $h$ is the embedding length. The embeddings are then passed into two decoders to compute the mean weight matrix $\rho_{\Omega^B}$, and its corresponding standard deviation $\sigma_{\Omega^B}$ (Fig. \ref{fig:policy}).

\vspace{-.2cm}
\subsection{Multi-head Attention (MHA) based decoding} \label{subsec:mha_decoding}
\vspace{-.2cm}
We initialize two MHA-based decoders, to compute the mean weights ($\rho_{\Omega^B}$) and the standard deviation ($\sigma_{\Omega^B}$). Each decoder takes in the task embeddings $F^{T}$ and the robot embeddings $F^{R}$ and outputs a matrix of size $N^{R} \times N^{T}$. The structure of the decoder is shown in Fig. \ref{fig:mha_decoder}. $F^{T}$ is used to compute the key $\mathcal{K}$ and value $\mathcal{V}$, while the a set of query $\mathcal{Q}$ is computed using $F^{R}$.
 The output of decoders are matrices for the mean ($\rho_{\Omega^B}$) and the standard deviation ($\sigma_{\Omega^B}$).

\vspace{-.1cm}
\subsection{BiGraph Weights Modeled as Probability Distributions}
\vspace{-.1cm}

The outputs from the decoder ($\rho_{\Omega^B}$ and $\sigma_{\Omega^B}$), which represent the matrices with the mean and the standard deviation for the bigraph weights, are then used to express $N^{R} \times N^{T}$ Lognormal probability distributions, from which the bigraph weights ($\Omega^B_{r,i}$) are drawn. To encourage exploration during training, the weights are sampled in an $\epsilon$-greedy fashion  ($\epsilon=0.2$); i.e., with $\epsilon$ probability the weight is randomly sampled from the corresponding distribution, and in the remaining cases, the mean value (from $\rho_{\Omega^B}$) is directly used. During testing, the mean value is always greedily used.

\vspace{-.1cm}
\subsection{Weighted Bigraph Construction}
\vspace{-.1cm}

Similar to \cite{ghassemi2021multi},  we omit the edges that represent an infeasible combination of robots and tasks, and those connecting tasks whose demand has been fully met or deadline has passed. The remaining edge weights are obtained from the computed weight matrix distributions. Once the weighted bigraph is constructed, we perform a maximum weight matching using the Hungarian Algorithm \cite{kuhn1955hungarian} to match robots to tasks. The deciding robot then broadcasts the task selection information to peer robots.

\vspace{-.1cm}
\subsection{BiG-CAM Policy Training Details}\label{subsec:training_details}
\vspace{-.1cm}

\textbf{\textit{Training Dataset}:} We train the BiG-CAM policy on an environment with 50 tasks, 6 robots, and a depot. Every episode is a different scenario characterized by the task location, demand, time deadline, and depot location. The locations of the tasks and the depot are randomly generated following a uniform distribution $\in[0,1]$ km. We consider every robot to have a constant speed of 0.01 km$/$sec. The demand for the tasks is an integer drawn randomly between 1 and 10 following a uniform distribution. The time deadline for the tasks is drawn from a uniform distribution $\in[165,550]$ seconds. The environment settings here are adopted from 
 \cite{paul2023efficient}, inspired by applications such as multi-robot disaster response and materials transport. For both TGE and RGE, we use the same settings for the GCAPS encoders as in \cite{paul2023efficient}. Eight attention heads are used in MHA-based decoders.

\textbf{Training Algorithm:}
In order to train the BiG-CAM policy, we use Proximal Policy Optimization (PPO) \cite{schulman2017proximal}. We allow 5 million steps, with a rollout buffer size of 50000, and a batch size of 10000. The simulation environment is developed in \textit{Python} as an \textit{Open AI gym} environment for ease of adoption. The training is performed on an Intel Xeon Gold 6330 CPU with 512GB RAM and an NVIDIA A100 GPU using PPO from \textit{Stable-Baselines3} \cite{stable-baselines3}. The testing is performed only using the CPU.
Here, we consider centralized training, where the collected experience of all the robots is used for training a single policy. Implementation is decentralized, where at any decision-making instance for a given robot, it executes the BiG-CAM policy to compute bigraph weights and runs maximal matching. 

\vspace{-.1cm}
\section{Experimental Evaluation}\label{sec:Experimental_Evaluations}
\vspace{-.1cm}

\subsection{Baseline Methods:}
\vspace{-.1cm}
Three different baselines are considered, a bigraph matching approach that uses expert-designed incentive function as edge weights, an RL-trained policy that directly provides task selections for robots, and a feasibility-preserving random walk approach. They are summarized below.

\textbf{\textit{Bi-Graph MRTA (BiG-MRTA)}}: BiG-MRTA~\cite{Ghassemi-bigmrta-mrs2019,ghassemi2021multi} uses a handcrafted incentive function to compute the weights of the edges connecting robots and tasks, based on which maximum matching is performed to decide task selection. This incentive for robot $r$ to choose task $i$ at a time $t$ is a product of two terms. The first term measures the remaining range if the robot chooses and completes task $i$ and returns to the depot. This term becomes zero if there's insufficient battery for the return. It represents the remaining potential for robot $r$ to perform additional tasks after task $i$. The second term is a negative exponential function of the time $t_i^r$ needed for robot $r$ to complete task $i$ if chosen next, i.e., before the deadline $\tau_i$. If task $i$ can't be completed by robot $r$ before the deadline, the edge weight ($\omega_{ri}$) becomes zero. Therefore, the weight $\omega_{ri}$ of a bigraph edge $(r,i)$ is expressed as:
\begin{equation}\small
\label{eq:edgeWeight}
\omega_{ri} = 
  \begin{cases}
    \max{(0,l_r)}\cdot \exp{\left(-\frac{t_{i}^r}{\alpha}\right)} & \text{if } t_{i}^r \leq \tau_i\\
    0  & \text{Otherwise}
  \end{cases}
  \vspace{-.2cm}
\end{equation}
where $l_r = \Delta_{r}^{t} - (d_{ri}+d_{i0})$, $\alpha$ is the max time (550 seconds),
 $d_{ri}$ is the distance between the current location/destination of robot $r$ and the location of task $i$, while $d_{i0}$ is the distance between the location of task $i$ and the depot.

\textbf{\textit{Capsule-Attention Mechanism (CapAM)}}: The CapAM policy network from \cite{Paul_ICRA} uses a GCAPS network as GNN and an MHA-based decoder to directly compute log probabilities for all available tasks given a state for the robot taking the decision. This method has demonstrated better performance compared to other encodings such as GCN and GAT \cite{paul2022scalable},  and standard RL \cite{paul2023efficient}. This policy has been trained with the same settings as that of BiG-CAM, (refer Section \ref{subsec:training_details}). Previous work on related problems have already enlightened on the optimality gap of of BiG-MRTA and CapAM w.r.t MINLP solutions \cite{ghassemi2021multi,paul2023efficient}, and hence expensive MINLP solutions are not generated here.

\textbf{\textit{Feasibility-preserving Random-Walk (Feas-Rnd)}:} is a myopic decision-making method that takes randomized but feasible actions, avoiding conflicts and satisfying other problem constraints. Feas-Rnd serves as a lower bound to compare the performance of the other methods. 


\vspace{-.1cm}
\subsection{Training Convergence:}
\vspace{-.1cm}
From the training history, it was observed that at the end of 5 million steps, BiG-CAM converged to an average total episodic reward of 0.53, compared to 0.51 achieved by CapAM. this training process for BiG-CAM took $\sim$21 hours while CapAM took $\sim$15 hours. This increase in training time is due to the computation overhead of the maximum weight matching algorithm, and the larger number of trainable weights for BiG-CAM compared to CapAM.

\vspace{-.1cm}
\subsection{Performance Analysis (\% Task completion)}
\vspace{-.1cm}
\begin{figure}[!ht]
    \centering
    \begin{subfigure}{1.0\linewidth}
    \centering
    \includegraphics[width=0.8\textwidth, height=.4\textwidth]{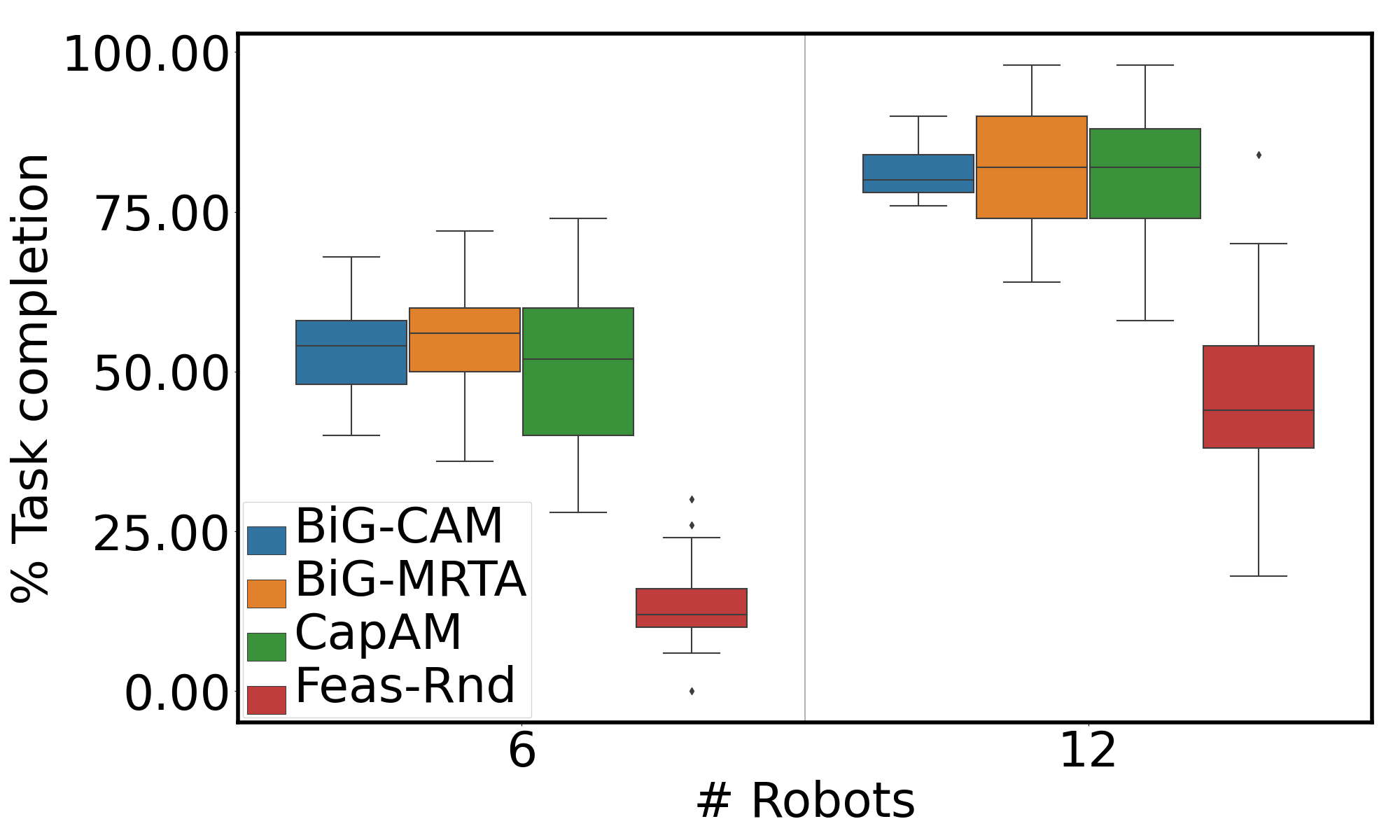}
    \caption{Scenarios with $N^{T}=50$ ($s_{t}$=1)}
    \label{fig:50_tasks_result}
    \end{subfigure}
    \hfill
    \begin{subfigure}{1.0\linewidth}
    \centering
    \includegraphics[width=0.8\textwidth, height=.4\textwidth]{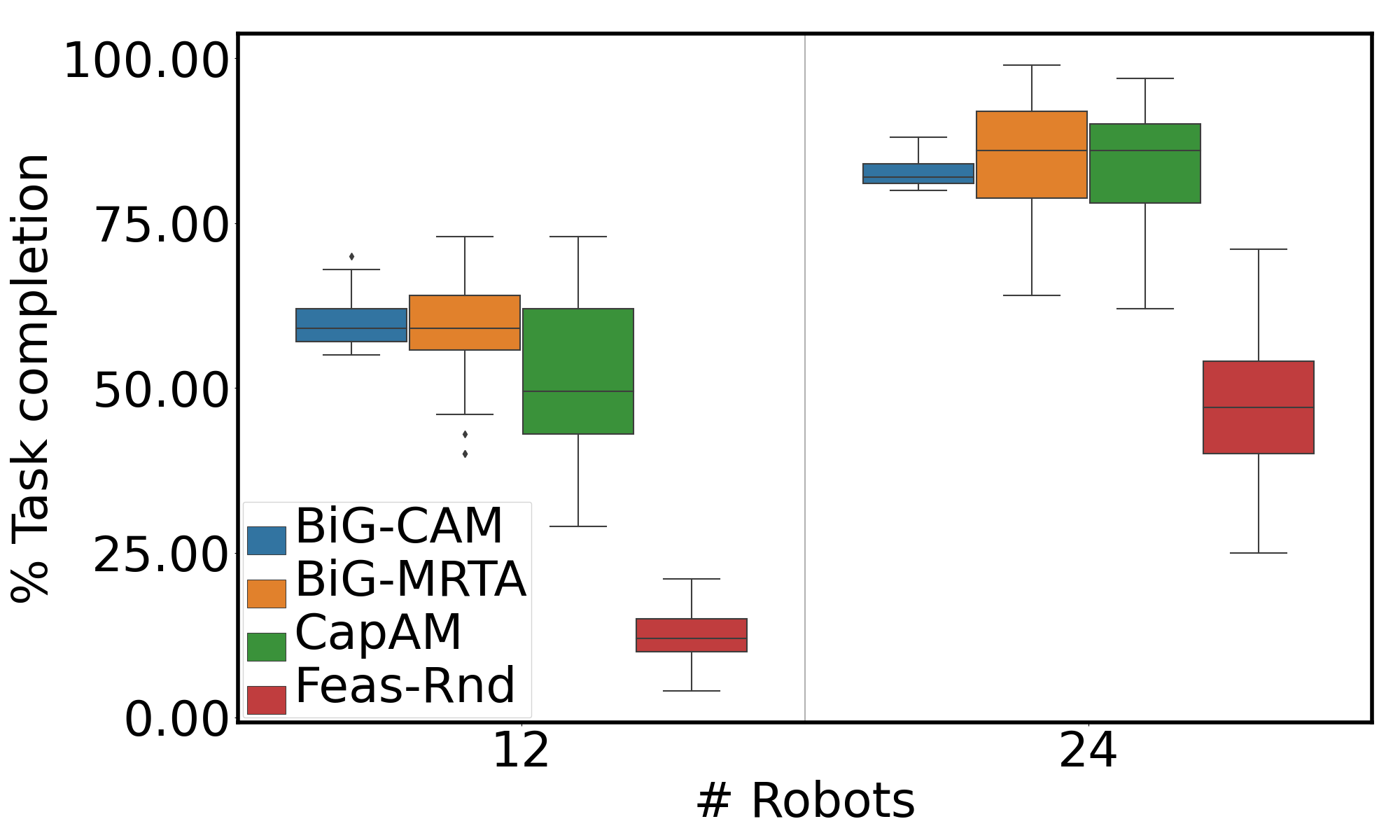}
    \caption{Scenarios with $N^{T}=100$ ($s_{t}$=2)}
    \label{fig:100_tasks_result}
    \end{subfigure}
    \begin{subfigure}{1.0\linewidth}
    \centering
    \includegraphics[width=0.8\textwidth, height=.4\textwidth]{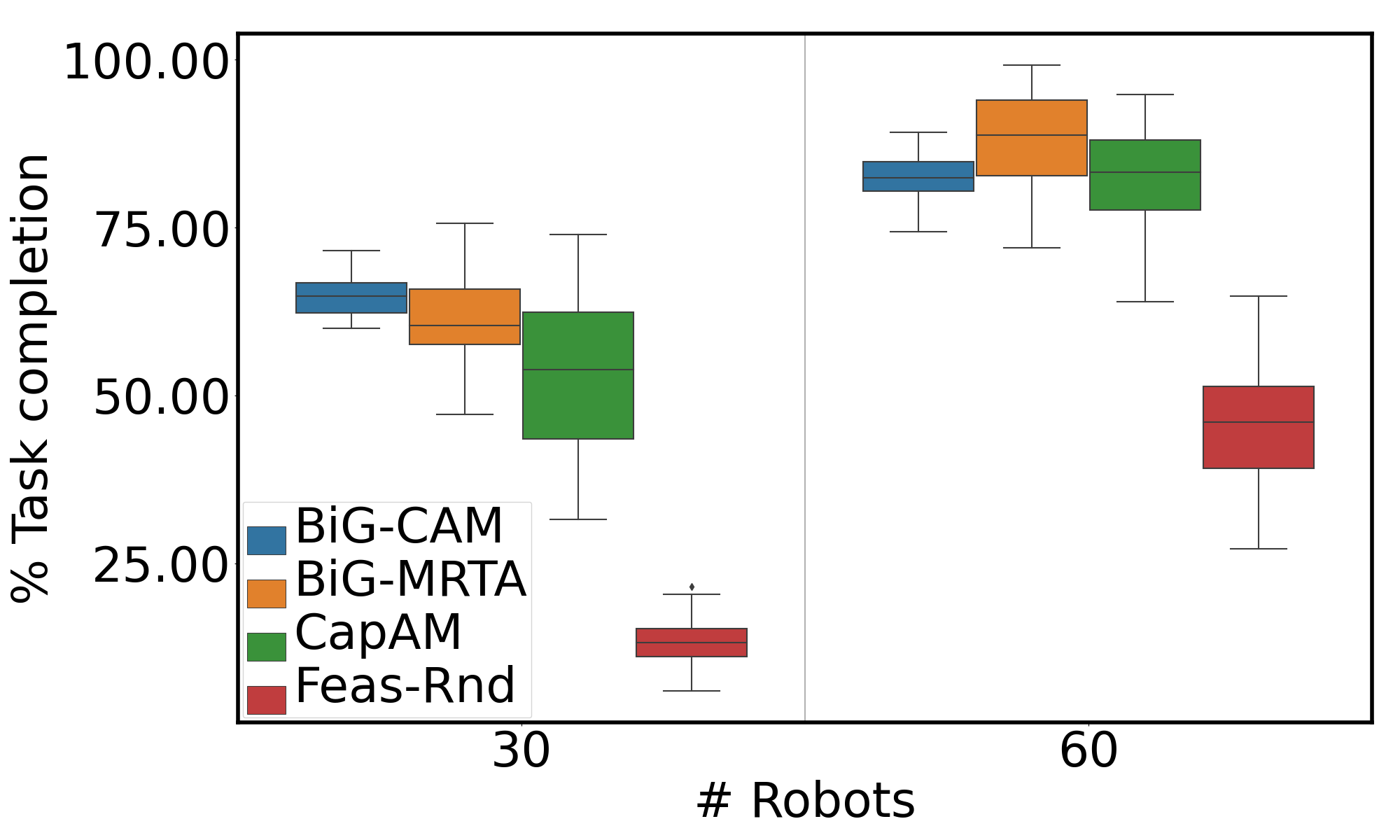}
    \caption{Scenarios with $N^{T}=250$ ($s_{t}$=5)}
    \label{fig:250_tasks_result}
    \end{subfigure}
    \hfill
    \begin{subfigure}{1.0\linewidth}
    \centering
    \includegraphics[width=0.8\textwidth, height=.4\textwidth]{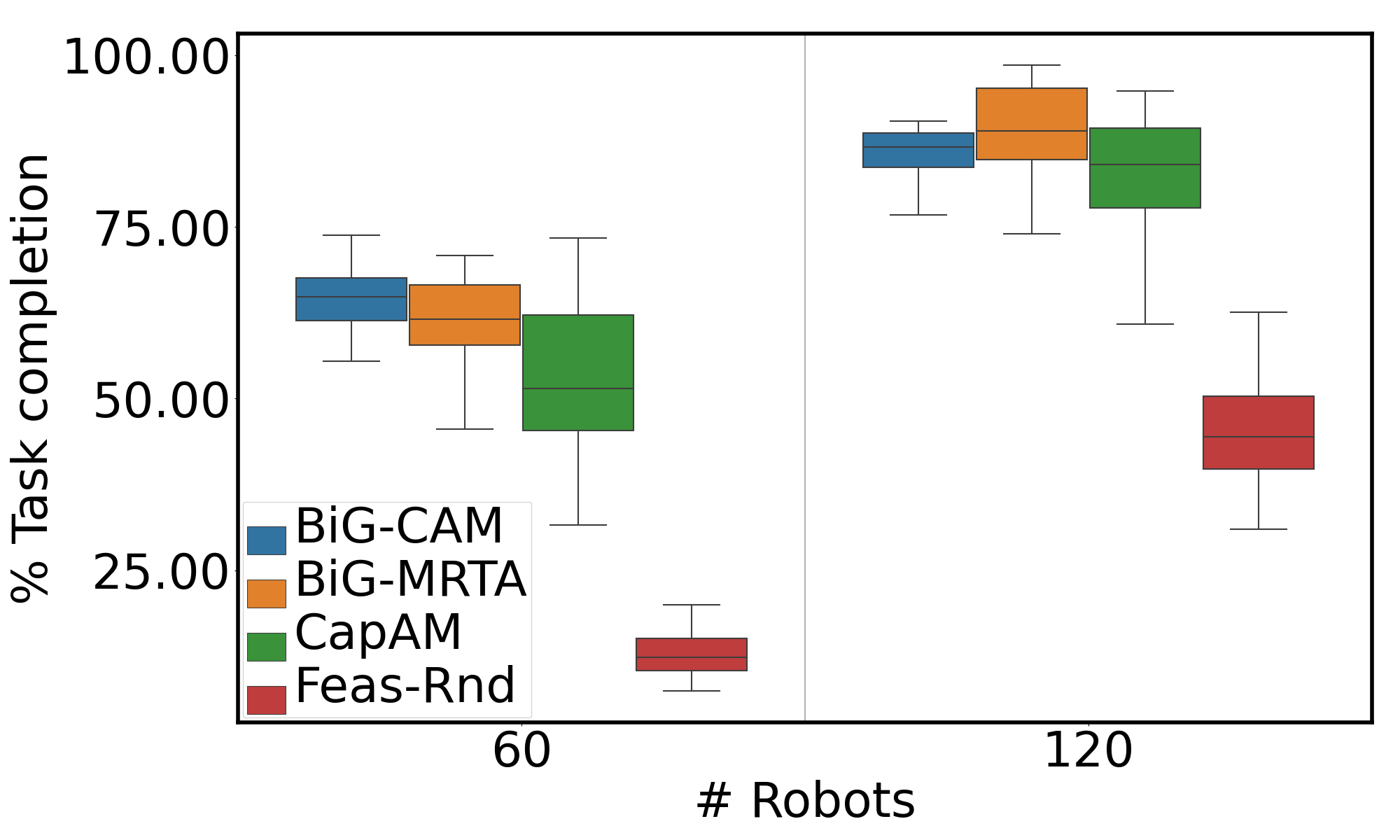}
    \caption{Scenarios with $N^{T}=500$ ($s_{t}$=10)}
    \label{fig:500_tasks_result}
    \end{subfigure}
    \caption{$\%$ task completion for all the methods. Left plots correspond to scenarios with $s_{r}=1$; right plots correspond to scenarios with $s_{r}=2$.}
    \label{fig:results}
    \vspace{-.6cm}
\end{figure}
In order to assess the performance of the learned policy on generalizability and scalability, we designed test scenarios with varying numbers of tasks $N^{T}$ and number of robots $N^{R}$. We generate different scenarios based on a set of scaling factors $S^{T}=\{1,2,5,10\}$ for tasks and $S^{R}=\{1,2\}$ for robots. For every combination of $s_{t} \in S^{T}$ and $s_{r} \in S^{R}$ we consider 100 scenarios with number of tasks $N^{T}=50\times s_{t}$, and number of robots $N^{R}=6\times s_{r}$ (note again, 50-task-6-robot scenarios were used in training). For example, for scenarios with $s_{t}$=2, and $s_{r}$=2, $N^{T}$=100, and $N^{R}$=24. 
We implemented the other baseline methods (BiG-MRTA, CapAM, and Feas-Rnd) on these same test scenarios for comparison in terms of $\%$ task completion metric. In order to confirm the significance of any performance difference between the methods, we performed the statistical t-test pairwise for different settings of $s_{t}$ and $s_{r}$, with the null hypothesis being the mean of the two sets of samples are the same. For BiG-CAM, for scenarios with $s_{r}$$>$1 and $s_{t}$$>$1, we shrink the size of the bigraph weight to 6$\times$50. The 50 tasks and 5 peer robots are chosen based on their proximity to the robot taking decision, thus keeping the computation time spent for the maximum weight matching relatively unchanged irrespective of the scenario size.

Across all testing scenarios, BiG-CAM, BiG-MRTA and CapAM provide clearly better task completion rates compared to Feas-Rnd. Comparing generalizability performance, i.e., in test scenarios of similar size as in training, we observe from Fig. \ref{fig:50_tasks_result} (left side) that BiG-CAM and BiG-MRTA exhibit comparable performance. BiG-MRTA has a slightly better median task completion \%, while BiG-CAM provides smaller variance. The t-test yields a p-value of 0.188, indicating no significant difference between BiG-CAM and BiG-MRTA for these scenarios. In contrast, when compared to CapAM, BiG-CAM has a slightly higher median (around 3\%) while maintaining a lower standard deviation. The p-value from the t-test is 0.03 ($<$0.05), indicating a significant difference between BiG-CAM and CapAM's performance. 

In scenarios with $s_{t}$=2 and fewer robots ($s_{r}$=1) (Fig. \ref{fig:100_tasks_result}, left side), BiG-CAM performs similarly to BiG-MRTA (p-value$>$0.05) while outperforming CapAM (p-value$<$0.05). However, for a larger number of robots $s_{t}$=2 (Fig. \ref{fig:100_tasks_result}, right side), BiG-CAM exhibits slightly inferior performance compared to both BiG-MRTA and CapAM (p-value$<$0.05 for both cases). For scenarios with $s_{t}$=5 and $10$ and for scenarios with a lower number of robots ($s_{r}$=1), BiG-CAM performs significantly better than both BiG-MRTA and CapAM (p-values $<$0.05), while for scenarios with a larger number of robots ($s_{r}$=2), BiG-CAM performs slightly poor compared to BiG-MRTA (p-value $<$0.05), and on par compared to CapAM for $s_{t}=5$ (p-value$>$0.05), and better than CapAM for $s_{t}$=10 (Figs. \ref{fig:250_tasks_result} and \ref{fig:500_tasks_result}).

BiG-CAM outperforms BiG-MRTA in scenarios with fewer robots ($s_{r}$=1), while BiG-MRTA excels in scenarios with more robots ($s_{r}$=2). The performance drop of BiG-CAM compared to BiG-MRTA could be partly because of the forced limiting of the bigraph to the size for which the policy model in BiG-CAM has been trained. 
Notably, BiG-CAM exhibits significantly lower variance across all scenarios compared to both BiG-MRTA and CapAM. BiG-CAM's standard deviation ranges from 0.064 ($s_{t}$=1, $s_{r}$=1) to 0.21 ($s_{t}$=2, $s_{r}$=2), while BiG-MRTA's standard deviation spans from 0.88 ($s_{t}$=1, $s_{r}$=2) to 0.57 ($s_{t}$=5, $s_{r}$=1). 

\textbf{Computing time analysis:} Computation time is assessed using two metrics: the average time for all decisions in an episode and the time for a single decision, as presented in Table \ref{table:comp_time}. CapAM which only includes a policy execution is faster than BiG-CAM and BiG-MRTa as expected. Now, between BiG-MRTA and BiG-CAM, scenarios with fewer tasks ($N^{T}$=50,100) favor BiG-MRTA, since it solely performs maximum weight matching. BiG-CAM needs to also perform policy execution to compute the bigraph weights, which adds to its computing time. However, in scenarios with more tasks ($N^{T}=250,500$), BiG-CAM is significantly faster than BiG-MRTA (up to 9 times). This is because BiG-CAM limits the bigraph size to 50-task/6-robot based on proximity, while BiG-MRTA considers the entire task/robot space leading to greater cost of the maximum matching process. 

\begin{table}[!ht]
\centering
\caption{Average total episodic decision computing time (with average computing time per decision) in seconds}
\label{table:comp_time}
\resizebox{.85\columnwidth}{!}{%
\begin{tabular}{|c|c|c|c|c|c|}
\hline
\textbf{$N^{T}$}                   & \textbf{$N^{R}$}  & \textbf{BiG-CAM} & \textbf{BiG-MRTA} & \textbf{CapAM} & \textbf{Feas-Rnd} \\ \hline
\multirow{2}{*}{50}  & 6   & 0.67 (0.007)                  &  0.28 (0.004)                 &  0.20 (0.003)              &           0.002 (2e-5)        \\ \cline{2-6} 
                     & 12  &  1.3 (0.007)                &   0.7 (0.006)                &    0.28 (0.002)            &  0.002 (1.6e-4)                 \\ \hline \hline
\multirow{2}{*}{100} & 12  &  2.26 (0.012)                &    1.6 (0.007)               &      0.64 (0.004)          &            0.003 (1.7e-5)       \\ \cline{2-6} 
                     & 24  &  5.47 (0.012)                & 4.7 (0.009)                  &    0.92 (0.004)            &        0.005 (1.8e-5)           \\ \hline \hline
\multirow{2}{*}{250} & 30  &  12.99 (0.02)                &    23.2 (0.07)               &      3.03 (0.009)          &           0.006 (1.8e-5)        \\ \cline{2-6} 
                     & 60  &    38.15 (0.04)              &       67.10 (0.2)            &     6.39 (0.009)           &            0.013 (1.8e-5)        \\ \hline \hline
\multirow{2}{*}{500} & 60  &  91.9 (0.09)                &   174.4 (0.40)                &       25.5 (0.03)         &         0.012 (1.9e-5)             \\ \cline{2-6} 
                     & 120 &      192.0 (0.11)            &       464.0 (0.90)            &     35.6 (0.03)           &           0.014 (1.9e-5)     \\ \hline
\end{tabular}%
}
\vspace{-0.4cm}
\end{table}

\vspace{-.1cm}
\subsection{Learned incentives vs expert-derived incentives model}\label{subsec:expert_vs_learned}
\begin{figure}[!ht]
\scriptsize
\vspace{-0.3cm}
    \centering
    \includegraphics[width=0.4\textwidth, height=0.2\textwidth]{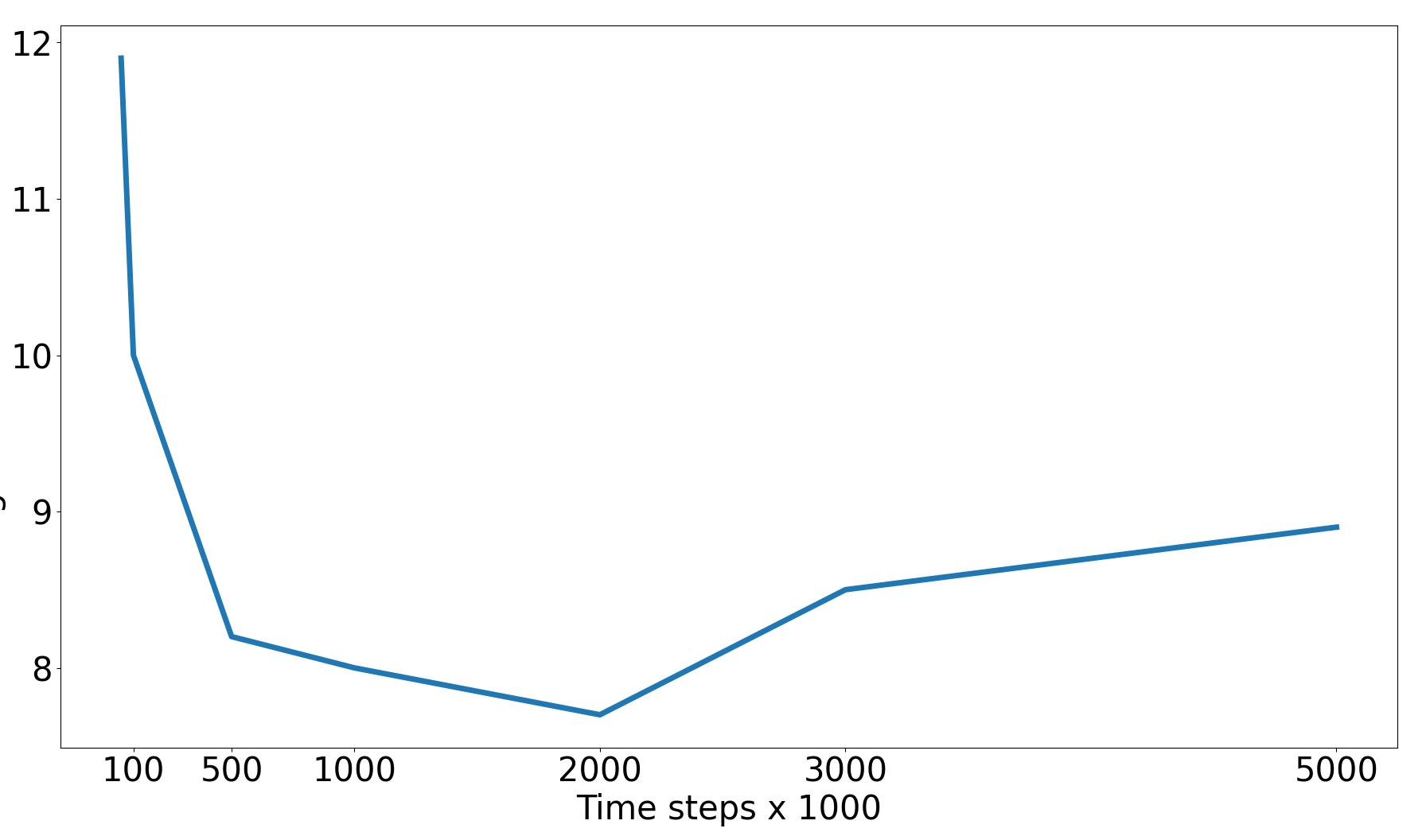}
    \caption{Comparison of Sinkhorn distance between the bigraph weight of $\mathbb{S}$.}
    \label{fig:sinkhorn}
    \vspace{-0.3cm}
\end{figure}

\vspace{-.1cm}
A key question is whether the learned policy in BiG-CAM produces incentives that are similar to or differ from those computed by the expert-specified incentive function in BiG-MRTA. To answer this question, we compute the weight matrix for a set of 1000 states, $\mathbb{S}$ = [$S_{1}, \dots S_{1000}$] in 50-task/6-robot scenarios, using the policy at different stages of learning, namely when the time steps are 50K, 100K, 500K, 1M, 2M, 3M, and 5M. Then we compute the average \textit{Sinkhorn} distance between the weight matrices derived from the learned policy and the corresponding weight matrix given by BiG-MRTA expert-specified incentive function, as shown in Fig. \ref{fig:sinkhorn}. We observe that the average \textit{Sinkhorn} distance between the weight matrices of the two methods for all the states in $\mathbb{S}$ decreases until 2M, and then increases. This observation shows that as the learning progresses the weight matrix initially gets more similar to that given by the expert-specified incentive function (in BiG-MRTA), but later on slightly deviates from the expert incentive function.

\vspace{-.4cm}
\section{Conclusions}\label{sec:Conclusion}
\vspace{-.2cm}
This paper proposed a graph RL approach called BiG-CAM to learn incentives or weights for a bigraph representation of candidate robot-task pairing in MRTA, which is then used by a maximum weight matching method to allocate tasks. We considered an MRTA collective transport (MRTA-CT) problem, which was formulated as an MDP with the state of the tasks and the robots expressed as graphs. The weights of the task/robot pairing bigraph are sampled from distributions computed by a policy network (BiG-CAM) that act on the state space graphs, comprises GNN encoders and MHA-based decoders, and trained using PPO. In testing, BiG-CAM demonstrated comparable or slightly better performance relative to BiG-MRTA (that instead uses an expert-crafted incentive to compute bigraph weights) for scenarios with lower number of robots, and comparable or slightly poorer median performance for scenarios with a larger number of robots. Compared to both BiG-MRTA and CapAM (purely GNN for MRTA), BiG-CAM demonstrated better robustness w.r.t. task completion rates. In the future, alleviating the limitation of fixing the size of the task and robot spaces during training of BiG-CAM could further improve its relative performance. Future systematic analysis is also needed to adapt the bigraph size based on the expected propagation of decision influence across the task/robot graphs to ensure reliable yet compute-efficient scalability of the underlying bigraph matching concept. 

\bibliographystyle{./IEEEtran}
\bibliography{./ICRA_2024}



\end{document}